\documentclass[journal]{IEEEtran}
\usepackage{graphicx}
\usepackage{amsmath}
\usepackage{amssymb}

\usepackage{hyperref}
\hypersetup{
    colorlinks=true,
    linkcolor=black,
    filecolor=blue,      
    urlcolor=blue,
    }

\usepackage{float} 

\begin{document}

\title{Quantum deep reinforcement learning for \\ humanoid robot navigation task}

\author{
    \IEEEauthorblockN{Romerik Lokossou\IEEEauthorrefmark{1}\thanks{The authors Birhanu Shimelis Girma and Romerik Lokossou contributed equally to this work. \\
    The repository for the implementation can be found on \href{https://github.com/romerik/Quantum_Deep_Reinforcement}{GitHub}}, Birhanu Shimelis Girma\IEEEauthorrefmark{1}, Ozan K. Tonguz\IEEEauthorrefmark{2}, Ahmed Biyabani\IEEEauthorrefmark{1}} \\
    \IEEEauthorblockA{\IEEEauthorrefmark{1}Carnegie Mellon University Africa\\
    \IEEEauthorrefmark{2}Carnegie Mellon University\\    
    \{rlokosso, bgirmash\}@alumni.cmu.edu, \{tonguz, ab3x\}@andrew.cmu.edu}
}

\maketitle
\begin{abstract}

Classical reinforcement learning (RL) methods often struggle in complex, high-dimensional environments because of their extensive parameter requirements and challenges posed by stochastic, non-deterministic settings. This study introduces quantum deep reinforcement learning (QDRL) to train humanoid agents efficiently. While previous quantum RL models focused on smaller environments, such as wheeled robots and robotic arms, our work pioneers the application of QDRL to humanoid robotics, specifically in environments with substantial observation and action spaces, such as MuJoCo's Humanoid-v4 and Walker2d-v4. Using parameterized quantum circuits, we explored a hybrid quantum-classical setup to directly navigate high-dimensional state spaces, bypassing traditional mapping and planning. By integrating quantum computing with deep RL, we aim to develop models that can efficiently learn complex navigation tasks in humanoid robots. We evaluated the performance of the Soft Actor-Critic (SAC) in classical RL against its quantum implementation. The results show that the quantum SAC achieves an 8\% higher average return (246.40) than the classical SAC (228.36) after 92\% fewer steps, highlighting the accelerated learning potential of quantum computing in RL tasks. 

\end{abstract}

\IEEEpeerreviewmaketitle
\section{Introduction}
Classical reinforcement learning (RL) methods have proven effective for a wide range of simple to moderately complex tasks. However, their application in high-dimensional and complex environments, such as those encountered in humanoid robot navigation, is often limited by the exponential growth in the number of parameters required for effective learning. These environments require agents to process large amounts of sensory input and perform intricate motor control, increasing the challenges of training, stability, and convergence speed \cite{sutton_reinforcement_nodate}.

Furthermore, classical RL methods may struggle with environments characterized by stochastic and non-deterministic elements, as these can lead to significant variance in training results. This variability often requires a larger number of training episodes to achieve a reliable performance, thus increasing the computational burden \cite{mnih_human-level_2015}.

To address these challenges, there has been growing interest in the use of quantum computing within the machine learning domain. Quantum computing offers potential breakthroughs in processing power and efficiency owing to its inherent properties of superposition, entanglement, and quantum interference\cite{biamonte_quantum_2017}. Quantum deep reinforcement learning (QDRL) has emerged as a promising hybrid approach that combines the strengths of quantum computing with the adaptability of deep reinforcement learning. This integration aims to reduce the parameter space and enhance the speed of convergence when learning complex tasks\cite{dunjko_machine_2017}.

Quantum RL models have shown promise in simplifying the complexities of RL problems. For example, quantum annealing and gate-based quantum computing have been applied to reduce the number of training steps required to learn optimal policies for navigation tasks. These quantum approaches exploit the ability to perform parallel computations in multiple states of a problem simultaneously, potentially leading to faster learning rates \cite{crawford_reinforcement_2019}.

This study builds on the foundational work in quantum RL, aiming to employ and expand these techniques for humanoid robot navigation, a task characterized by high-dimensional state spaces and the need for dynamic balance and coordination. Our approach seeks to train a humanoid agent not only to walk but also to optimize its walking speed while maintaining stability, utilizing a hybrid quantum-inspired framework. We conducted experiments with Soft Actor-Critic (SAC) and different algorithms for the robot navigation task in a simulated environment to evaluate their performance. This study contributes to the understanding and application of quantum computing in complex real-world tasks.

\section{Related Work}

Quantum Reinforcement Learning (QRL) has emerged as a promising area in robotics and autonomous systems. The progression from foundational theories to complex applications has significantly advanced the field, with each study building upon the previous one to address specific limitations and explore new capabilities.

An early study by Dong et al. (2012) laid the groundwork by demonstrating the potential of a quantum-inspired reinforcement learning (QiRL) algorithm in real and simulated mobile-robot navigation scenarios. This study compared the QiRL approach with classical reinforcement learning methods, demonstrating its effectiveness in managing different learning rates and initial states. However, it highlighted the need for further exploration of the application of QiRL in more varied and complex environments, suggesting that its robustness could be enhanced with additional empirical evidence \cite{dong_robust_2012}.

Building on the concept of quantum enhancement in learning algorithms, Dragan et al. (2022) focused on a hybrid quantum-classical model to address the challenges in a stochastic frozen lake environment. Their research evaluated different quantum architectures based on their entanglement capability, expressibility, and information density, demonstrating improvements in performance, parameter reduction, and solution times. Despite these advancements, the study primarily explored simplified gym environments, indicating a gap in the application of these models to more dynamic and unpredictable real-world tasks \cite{icaart23}.

Further specializing in the application of quantum computing in robotics, Heimann et al. applied Quantum Deep Reinforcement Learning to teach navigation tasks to a wheeled robot\cite{heimann_quantum_2022}. Utilizing a parameterized quantum circuit in a hybrid setup, their approach achieved higher performance than classical methods in complex indoor environments. Although this marked significant progress, the study highlighted the limitations of traditional navigation methods, such as the reliance on internal maps and pre-planned routes, suggesting a gap in learning directly from environmental interactions without predefined assumptions \cite{10568148}.

Expanding the application of QRL to social scenarios, Samsani et al. (2021) explored the use of deep reinforcement learning to guide social robots in crowded environments. Their model successfully predicted human behavior in real time and navigated safely in multi-agent scenarios using advanced methods such as CADRL, LSTM-RL, and SARL. However, this study focused mainly on safety, indicating a potential gap in optimizing other interactive behaviors and adapting to rapidly changing human dynamics, which are crucial for broader applications in real-world social settings \cite{samsani_socially_2021}.

Alberto Acuto et al., explored the application of Quantum Computing to enhance the Soft Actor-Critic (SAC), a state-of-the-art Reinforcement Learning technique, for controlling a robotic arm in simulated environments. The primary focus is on employing a variational quantum soft actor–critic approach, where digital simulations of quantum circuits are used to potentially reduce the required number of parameters significantly, thereby improving the efficiency of model training. This quantum advantage is demonstrated by the decreased parameter requirements for achieving satisfactory performance compared with classical algorithms. However, the study highlights limitations, such as constrained real-world applicability due to experiments being confined to simulations and the need for stronger evidence of quantum advantages across broader scenarios.  Moreover, their work primarily applies to robotic arms, which typically have lower observation and action space. In contrast, our research extends this approach to humanoid robots, where tasks are inherently more complex, thus presenting a novel and challenging frontier for quantum-enhanced reinforcement learning techniques. \cite{Acuto:2022ozp}.

Recently, Sinha et al. (2025) introduced Nav-Q, the first quantum-supported deep reinforcement learning algorithm specifically designed for the collision-free navigation (CFN) of self-driving cars. Similar to our work on humanoid robots, Nav-Q tackles a complex real-world challenge involving high-dimensional observation and action spaces, although in a different domain. Their approach utilizes an actor-critic framework, where the critic is implemented using a hybrid quantum-classical algorithm, paralleling our QDRL architecture but with a key distinction. While we implemented a quantum SAC algorithm with PQCs in both humanoid and walker environments, Nav-Q incorporates quantum computing only in the critic component. Furthermore, they introduced a technique called qubit-independent data encoding and processing (QIDEP), which extends the data re-uploading strategy we also employ, but with a specialized focus on encoding high-dimensional vectors using an arbitrary number of qubits\cite{sinha_nav-q_2025}. A notable innovation in their approach is that Nav-Q leverages quantum computation only during training without requiring onboard quantum hardware during testing, making it practical for real-world deployment. The authors evaluated Nav-Q using the CARLA driving simulator, demonstrating that their quantum implementation outperformed its classical counterpart in terms of training stability, exhibiting reduced sensitivity to different weight initializations and slightly higher average cumulative rewards. Their analysis of the effective dimension revealed that incorporating quantum components resulted in models with greater descriptive power than classical baselines, a finding that reinforces our observations regarding parameter efficiency in quantum implementations. This work, alongside our research, represents significant progress in applying quantum reinforcement learning to complex navigation tasks in dynamic, partially observable environments with high-dimensional state spaces, moving beyond the simplified OpenAI Gym environments used in previous quantum RL studies.

\section{Problem Statement}

Quantum algorithms have shown potential in Reinforcement Learning (RL) models, typically in two approaches: replacing the agent with quantum-inspired algorithms \cite{bhandari_application_2023} or substituting classical neural networks with quantum circuits \cite{heimann_quantum_2022}. However, existing applications primarily focus on simplified or specific tasks, such as navigation in wheeled robots or control of robotic arms. This study aims to extend these approaches to the more complex domain of humanoid robotics, which presents unique challenges in terms of action and observation space dimensions.
This study tested a deep quantum-inspired RL algorithm. Specifically, we employed the Soft Actor-Critic (SAC) algorithm to explore its effectiveness in navigating the complex decision spaces inherent to humanoid robots. We will not only utilize standard benchmark quantum circuits but also experiment with parameterized quantum circuits. Both the quantum-inspired and classical approaches are compared to evaluate \hyphenation{op-tical net-works semi-conduc-tor}performance improvements, particularly in terms of learning efficiency and computational resource requirements.

\section{Approach}
\subsection{Learning Environment Setup: Walker2D gym environment } 

    The Walker2D gym environment, provided by the OpenAI Gym toolkit, serves as a benchmark for learning locomotion tasks in simulated bipedal walkers. Unlike simpler environments, Walker2D presents a significant challenge because of its vast observation and action spaces, both of which are continuous. The observation space comprises 17 dimensions, including joint angles, velocities, and body orientation, which provide detailed information regarding the state of the robot. In contrast, the action space consisted of six continuous actions representing the torque commands applied to the robot joints. The complexity of the observation and action spaces in Walker2D makes it a suitable test bed for evaluating reinforcement learning algorithms for high-dimensional continuous control tasks.

    In the context of our research, which focuses on applying quantum deep reinforcement learning (QDRL) to humanoid robotics, the Walker2D environment is particularly relevant. The large observation space, which consists of 17 dimensions, reflects the rich sensory information available to a humanoid agent when navigating its surroundings. Similarly, the 6-dimensional action space offers a wide range of possibilities for controlling the agent's movements, allowing for nuanced responses to environmental cues. By addressing the challenges posed by continuous observation and action spaces in Walker2D, our work aims to demonstrate the effectiveness of quantum-inspired approaches in addressing complex navigation tasks for humanoid robots.

    \begin{figure}[htbp]
    \includegraphics[width=0.3\textwidth]{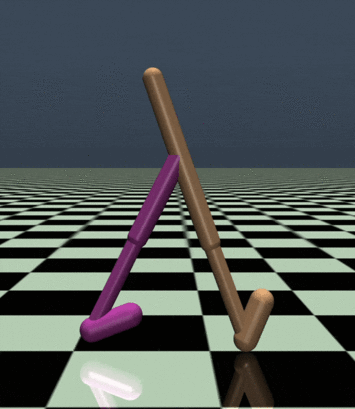}
    \caption{Walker2D-v4 Gym environment} \label{fig_walker} 
    \end{figure}
   
\subsection{Relevance of Soft Actor-Critic in Walker2D Gym Environment}

The Soft Actor-Critic (SAC) emerged as a key choice in our comparative study between classical deep reinforcement learning (DRL) and quantum computing approaches within the Walker2D Gym environment. The selection of SAC as our focal algorithm was due to its robustness in addressing the inherent challenges in deep reinforcement learning, making it an ideal candidate for exploration in different computational paradigms.

In the realm of classical DRL, SAC's adoption of SAC is justified by its ability to confront the fundamental limitations encountered in traditional actor-critic methods. Notably, SAC's integration of a maximum entropy framework in SAC facilitates efficient exploration in high-dimensional state and action spaces, which is characteristic of environments such as Humanoid Mujoco. By striking a balance between long-term rewards and entropy, the SAC offers a principled approach to navigating complex environments, minimizing the need for overly rigid hyperparameter tuning. This adaptability is crucial to ensure stable and effective learning, particularly in scenarios with sparse rewards or intricate dynamics.

Furthermore, SAC's architectural design of the SAC, which leverages multiple neural networks for actors, value estimation, and critics, enhances its stability and resilience in challenging environments. The incorporation of twin critics and a target value function mitigates the issues of instability of the value function, contributing to more reliable and consistent learning outcomes. Additionally, SAC's probabilistic policy modeling of SAC introduces stochasticity into the decision-making process, enabling effective exploration while maintaining a principled approach to policy optimization.

Our research extends beyond the confines of classical DRL, venturing into the realm of quantum computing to explore SAC's performance of SAC in a novel computational paradigm. By leveraging the unique properties of quantum computation, we seek to elucidate the potential advantages of quantum reinforcement learning in complex tasks, ultimately aiming to establish quantum supremacy in Humanoid Mujoco environments.

\subsection{Variational Quantum Soft Actor-Critic for Deep Reinforcement Learning}

A Parameterized Quantum Circuit (PQC) operates by applying a sequence of quantum gates with adjustable parameters, which can be tuned during the learning process, similar to the optimization of weights in neural networks. These parameters are optimized through classical computation to achieve a desired quantum state output or to solve a specific problem, making PQC a hybrid quantum-classical approach often used in variational algorithms like the Variational Quantum Eigensolver (VQE) and the Quantum Approximate Optimization Algorithm (QAOA).

Building on the concept of PQCs, we implemented a Variational Quantum Soft Actor-Critic (QuantumSAC) approach that combines the advantages of quantum computing with the robustness of the Soft Actor-Critic (SAC) algorithm. This method is particularly suitable for the high-dimensional observation and action spaces encountered in humanoid robot navigation tasks.

\begin{figure}[htbp]
\centering
\includegraphics[width=0.45\textwidth]{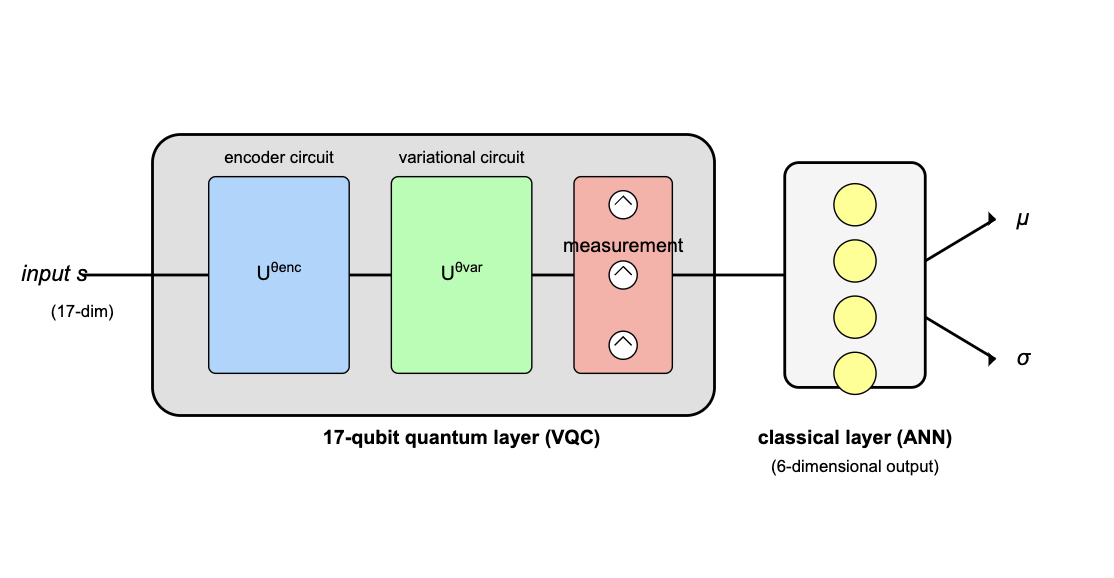}
\caption{The architecture of a hybrid quantum-classical policy network for QuantumSAC. The input state $s$ with 17 dimensions is processed by the encoder circuit $U_{\theta enc}$ and variational circuit $U_{\theta var}$ in a 17-qubit quantum layer. The measurement results are then passed to a classical neural network layer which outputs the mean $\mu$ and standard deviation $\sigma$ of the 6-dimensional action distribution representing joint torques.} \label{fig_quantum_sac} 
\end{figure}

In our work, we created a deep learning model that embeds a parameterized quantum circuit and trained it on the \textbf{Mujoco Walker2D-v4}, which is a less complex environment than the \textbf{Mujoco Humanoid-v4}, because the \textbf{Humanoid} requires too much computation owing to the complexity of the environment, and the current quantum computing systems that exist support only a few qubits.

\begin{figure}[htbp]
\includegraphics[width=0.5\textwidth]{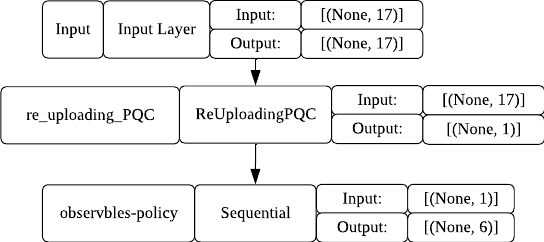}
\caption{Quantum deep learning model with parametrized quantum circuit} \label{fig_model} 
\end{figure}

Our hybrid quantum-classical policy network architecture, illustrated in Figure \ref{fig_quantum_sac}, includes the following:

\begin{enumerate}
    \item \textbf{Input Layer:} Takes a 17-dimensional input vector (corresponding to the state features in the MuJoCo Humanoid environment).
    
    \item \textbf{Re-uploading PQC (Parameterized Quantum Circuit):} A quantum layer with 17 qubits that repeatedly applies a parameterized quantum circuit on the input data. We specifically utilize a data re-uploading VQC approach where the input data is encoded multiple times throughout the circuit, significantly enhancing the expressivity of the quantum circuit. The circuit takes the 17-dimensional input and produces measurements that are then processed classically.
    
    \item \textbf{Observables-Policy:} A classical sequential layer that processes the outputs from the quantum layer and generates a 6-dimensional output vector representing the means $\mu$ and standard deviations $\sigma$ of the action distribution. In our case, this corresponds to the values of the different torques we are applying to the robot joints to make it move.
\end{enumerate}

The QuantumSAC algorithm follows the maximum-entropy reinforcement learning framework and optimizes both the expected return and policy entropy:

\begin{equation}
J(\theta) = \mathrm{E}_{S\sim D,\epsilon\sim N(0,1)}\left[\min_{i=1,2} Q_{\phi_i}(S, \tilde{A}_\theta) - \alpha \log(\pi_\theta(\tilde{A}_\theta|S))\right]
\end{equation}

where $\tilde{A}_\theta$ represents the actions sampled using the reparameterization trick, and $\alpha$ is the temperature parameter controlling the importance of entropy.

A key advantage of our QuantumSAC implementation is its parameter efficiency. Our 2-layer data re-uploading quantum circuit achieves comparable performance to classical networks with significantly fewer parameters (41 learnable parameters versus 1,250 parameters in the equivalent classical network). This parameter efficiency is particularly valuable in high-dimensional humanoid robot control, where classical networks often require extensive parameterization.

For evaluation, we compared the performance of the soft actor–critic (SAC) in classical reinforcement learning against its quantum implementation. The results show that quantum SAC achieves an 8\% higher average return (246.40) than classical SAC (228.36) after 92\% fewer steps, highlighting quantum computing's accelerated learning potential in reinforcement learning tasks.

\section{Result and Comparison}
\subsection{Experimental Setup and Evaluation Criteria}

To evaluate the performance of the proposed Quantum Deep Reinforcement Learning (QDRL) approach, we conducted experiments in the MuJoCo Walker2d-v4 environment. The experiments aimed to compare the efficiency and effectiveness of the soft actor–critic (SAC) algorithm in its classical form against its quantum implementation.

The primary metrics used to evaluate the performance included the average return, defined as the cumulative reward achieved by the agent over an episode; the return over episode, which is the reward obtained in each individual episode; the average return over episode, representing the mean reward obtained over multiple episodes; the speed of running over episode, indicating the speed at which the agent completes tasks within each episode; and the cumulative steps over episode, representing the total number of steps taken by the agent over the episodes.

These metrics allowed us to comprehensively assess the performance, learning efficiency, and stability of the classical and quantum SAC algorithms. Specifically, our objective was to compare the speed of quantum SAC relative to classical SAC in terms of achieving high rewards in a shorter amount of time and experimenting more efficiently with the environment. These comparisons provide insights into the accelerated learning capabilities and overall effectiveness of the quantum SAC algorithm.

\subsection{Results}
\subsubsection{Return Over Episodes}
\begin{figure}[htbp]
\includegraphics[width=0.48\textwidth]{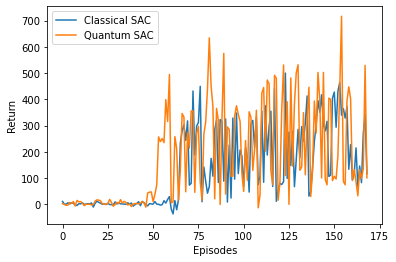}
\caption{Return of Classical SAC versus Quantum SAC in the Walker2d-v4 environment.} \label{return} 
\end{figure}

Figure \ref{return} illustrates the returns obtained by the Classical SAC and Quantum SAC algorithms over the training period. The x-axis represents the number of episodes, and the y-axis indicates the return received by each agent per episode.

As depicted in Figure \ref{return}, both the classical and quantum SAC algorithms exhibited an increase in return over time, indicating learning and adaptation to the environment. However, the Quantum SAC demonstrates a consistently higher average return than the Classical SAC throughout the training period. The Quantum SAC achieves peak performance faster and with fewer fluctuations in returns, suggesting a more stable and efficient learning process.

Additionally, Quantum SAC reaches higher returns more quickly, showcasing its ability to achieve significant rewards in a shorter time. This accelerated performance is indicative of the enhanced exploration and exploitation capabilities provided by quantum implementation.

\subsubsection{Average Return Over Episodes}

To further analyze the performance, Figure 2 presents the average return over episodes for both the classic SAC and quantum SAC algorithms. The x-axis represents the number of training steps, and the y-axis indicates the average return over episodes.

\begin{figure}[htbp]
\includegraphics[width=0.48\textwidth]{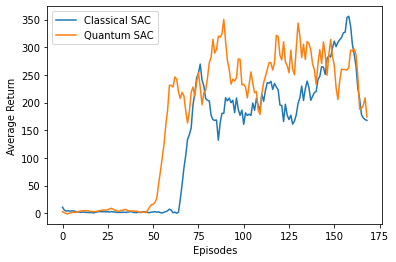}
\caption{Average return over episodes for Classical SAC versus Quantum SAC in the Walker2d-v4 environment.} \label{avg-return} 
\end{figure}

Figure \ref{avg-return} reveals that the Quantum SAC algorithm achieves higher average returns over time and does so consistently over multiple episodes. The Quantum SAC shows a more rapid increase in the average return early in the training process and maintains higher performance levels than the Classical SAC. This consistency underscores the efficiency of the Quantum SAC in learning and adapting to the environment more quickly.

The results of both figures reinforce the advantages of the Quantum SAC algorithm in terms of speed, stability, and overall learning efficiency. The ability of Quantum SAC to achieve significant rewards faster and with greater consistency highlights its potential for applications requiring rapid learning in complex environments.

\subsubsection{Steps per second over Average return}

\begin{figure}[htbp]
\includegraphics[width=0.5\textwidth]{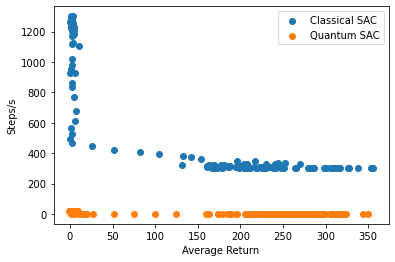}
\caption{Steps per second over average return for Classical SAC versus Quantum SAC in the Walker2d-v4 environment.} \label{speed} 
\end{figure}

Figure \ref{speed} shows the performance comparison between the classical SAC and quantum SAC algorithms during the training period. The x-axis represents the average return achieved by each agent, whereas the y-axis indicates the number of steps per second (SPS) required to reach that level of performance.

A performance comparison between classical and quantum implementations of the soft actor–critic (SAC) algorithm for the Walker2D environment revealed striking differences in learning efficiency and computational requirements. As illustrated in Figure \ref{speed}, the classical SAC initially operates at a high rate of approximately 1200 steps per second (SPS), gradually decreasing to 300-400 SPS as training progresses. This high initial step rate corresponds to low average returns, which improve over time, eventually reaching a maximum of approximately 350. In contrast, the quantum SAC maintained a consistently low SPS, close to 0, throughout the entire training process, but achieved higher average returns more rapidly, peaking at approximately 250.
The ability of the quantum SAC to attain comparable or superior performance with significantly fewer computational steps underscores its enhanced learning efficiency. This remarkable discrepancy in performance can be attributed to the unique properties of quantum computers. The principle of quantum superposition probably enables the algorithm to simultaneously explore multiple states within the environment, leading to a more comprehensive sampling of the state space at each step. Furthermore, quantum entanglement may allow the algorithm to capture and process complex correlations within the environment more effectively than classical counterparts. The expressive power of quantum circuits in representing the policy and value functions could also contribute to this enhanced performance, allowing for a more nuanced and efficient modeling of the dynamics of the environment. These quantum advantages manifest as a more stable learning curve for the quantum SAC, with less fluctuation in both the SPS and average return, suggesting a more robust and consistent learning process. The classical SAC requires a high number of initial steps, followed by a gradual reduction, indicating a less efficient exploration and exploitation balance than the quantum implementation. These findings align with our hypothesis that quantum deep reinforcement learning can offer significant advantages in complex environments such as Walker2D, demonstrating the potential of quantum computing to revolutionize reinforcement learning in high-dimensional and challenging tasks.

\begin{table}[htbp]
\centering
\begin{tabular}{|l|c|c|}
\hline
\textbf{Metric} & \textbf{Classical SAC} & \textbf{Quantum SAC} \\ \hline
Total Episodes       & 192.000000   & 169.000000   \\ \hline
Average Return       & 135.070156   & 172.661657   \\ \hline
Max Return           & 500.520000  & 716.710000   \\ \hline

\end{tabular}
\caption{Metric Comparison between Classical SAC and Quantum SAC}
\label{tab:sac_comparison}
\end{table}

Table 1 shows that quantum SAC achieves higher returns by leveraging quantum-inspired techniques that enhance its learning process, making it both more efficient and effective than the Classical SAC. Despite being trained over fewer episodes, the Quantum SAC consistently outperforms the Classical SAC by utilizing advanced exploration methods, such as quantum probabilistic sampling, which allows it to discover more optimal strategies and avoid local optima. The quantum-inspired SAC balances exploration and exploitation, enabling the Quantum SAC to make more impactful learning steps and extract more valuable information from each interaction with the environment. Additionally, its robustness to noise and tendency to search for global optima contribute to its ability to achieve higher maximum returns. These mechanisms allow the Quantum SAC to deliver superior performance, demonstrating its supremacy in reinforcement learning tasks.

\section{Conclusion}

The experimental results provide a clear comparison between the classical soft actor–critic (SAC) algorithm and its Quantum SAC implementation in the MuJoCo Walker2d-v4 environment.

Quantum SAC consistently achieves higher and more stable returns over episodes than Classical SAC, reaching peak performance more rapidly. It also maintains higher average returns throughout the training period, underscoring its efficiency in learning and adapting.

With respect to speed, the Classical SAC initially performed tasks at a higher rate but showed significant performance degradation over time. In contrast, the quantum SAC maintained a consistent speed, indicating a more controlled and steady learning process. The cumulative step analysis revealed that the Quantum SAC took more steps more rapidly, suggesting more active exploration and extensive interaction with the environment.

In conclusion, the quantum SAC outperformed the classical SAC in terms of learning efficiency, stability, and overall performance. Its ability to achieve significant rewards quickly and consistently highlights its potential for applications that require rapid and reliable learning in complex environments. These findings indicate that quantum enhancements in reinforcement learning can offer substantial benefits over classical approaches, encouraging further research and development of quantum-based AI technologies.

\section{Future Work}

Future research should evaluate the scalability of the Quantum SAC across different environments and tasks and investigate its performance in more complex and diverse scenarios. Exploring hybrid reinforcement learning models that combine classical and quantum techniques could leverage the strengths of both approaches and potentially offer improved performance.

Continued advances in quantum hardware will play a crucial role in the practical implementation of quantum reinforcement learning. Research should keep up with hardware developments to optimize the algorithms for emerging quantum processors. Further optimization of Quantum SAC algorithms can enhance their efficiency and effectiveness, including refining quantum circuits and reducing computational overhead.

Applying quantum SAC to real-world problems, such as robotics and autonomous systems, will provide valuable insights into its practical utility and impact in real-world applications. Collaboration with industry partners could facilitate the deployment of quantum reinforcement learning in practical settings. Deeper theoretical investigations into the underlying principles of quantum reinforcement learning could uncover new mechanisms that drive its superior performance, leading to the development of more powerful quantum algorithms.

\bibliographystyle{IEEEtran}

\bibliography{references}

\end{document}